\documentclass[runningheads]{llncs}
\usepackage{amsmath,amssymb,amsfonts}
\usepackage{graphicx}
\begin{document}
\title{Image Forgery Detection with Interpretability}
%
%
\author{Ankit Katiyar\inst{1} \and Dr. Arnav Bhavsar\inst{2}}
%
\authorrunning{F. Author et al.}
%
\institute{Indian Institute of Technology Mandi, HP, INDIA \\
\email{katiyar786ankit@gmail.com}\\ \and
Indian Institute of Technology Mandi, HP, INDIA\\
\email{arnav@iitmandi.ac.in}}
\maketitle              
\begin{abstract}
In this work, we present a learning based method
focusing on the convolutional neural network (CNN) architecture to detect these forgeries. We consider the detection of both copy-move forgeries and inpainting based forgeries. For these, we synthesize our own large dataset.  
In addition to classification, the focus is also on interpretability of the forgery detection. As the CNN classification yields the image-level label, it is important to understand if forged region has indeed contributed to the classification. For this purpose, we demonstrate using the Grad-CAM heatmap, that in various correctly classified examples, that the forged region is indeed the region contributing to the classification. Interestingly, this is also applicable for small forged regions, as is depicted in our results. Such an analysis can also help in establishing the reliability of the classification. 

\end{abstract}
\section{Introduction}
The fast development of digital image processing technology, advancement of image editing tools like and the omnipresence of the digital camera, makes it quite easy to edit and tamper digital images. These tampering and editing images looks so natural that humans often cannot tell whether image is forged or real. While in many cases image editing is used for well-intended entertainment or artistic purposes, in some cases, such forgeries can also be misused. This has led to a rise in the cases of image forgery in several fields relating to surveillance, crime, and is an important aspect of modern forensic investigations.

There are many ways to forge images such as inpainting, splicing, copy-move forgeries etc. In splicing a part of an image is copied and pasted on other image, whereas in copy-move forgery (CMF) a portion of an image is copied and pasted within the same image. In both cases, before pasting a image part, one can also carry out various image processing operations on the copied part such as rotation, scaling, blurring, colour variations etc.  In image inpainting an object is removed and removed portions are filled using the background neighbouring colour and texture using an inpainting algorithm, such that the inpainted region looks quite natural. In this work we mainly
focus on copy move and inpainting forgery.

Image forgery detection methods are generally classified into two categories, the active detection methods and the passive methods\cite{Sharma2016ImageFA}. The active methods are generally based on digital signatures or watermarks. The major drawback of active image authentication is, for verification of the authenticity of an image, a watermark or a digital signature need to be embedded into an image at the time of capture or immediately after the image is captured.

Passive forgery detection is an alternative to active authentication which requires no active information available for the purpose of authentication. Traditionally, these techniques involved detecting forgeries by analyzing the low-level image pixel statistics or geometrical relations among the objects. 

In more recent times, the passive methods involve deep learning based approaches, wherein a network can learn abstract discriminating features among the forged and real images. In this work we focus on the learning based paradigm, and contribute as follows: 

\begin{itemize}
    \item We first use an in-house convolutional neural network (CNN) to classify forged and real images, separately for copy-move forgery and inpainting forgeries. We then use it for an integrated forgery detection, without much drop in performance. 
    \item Further, via Gradient Class Activation Map (Grad-CAM) we interpret the forgery detection by this network, and show that the heat maps correctly detect the locations of forgeries, even in cases of relatively small forged regions. We believe that this analysis is important as it makes the forgery detection approach more reliable.
\end{itemize}

\section{Related Work}

Below, we briefly discuss some earlier passive image forgery methods including traditional paradigms and a few contemporary learning based methods. 

\subsection{Format based techniques}
Format based techniques mainly work on specific image formats, in which JPEG format is most popular. Statistical correlation at a block level introduced by specific lossy compression schemes, has been shown to be helpful for image forgery detection. Some examples of such techniques are based on JPEG quantization \cite{Farid2008DigitalIB}, Double JPEG \cite{Luks2003EstimationOP} and JPEG blocking \cite{4217384}. 

For instance, in\cite{4217384}, the authors characterize the blocking artifacts using pixel value differences within and across block boundaries. These differences tend to be smaller within blocks than across blocks. When an image is cropped and re-compressed, a new set of blocking artifacts may be introduced. 
Within- and across-block pixel value differences are computed from 4-pixel neighborhoods that are spatially offset from each other by a fixed amount, where one neighborhood lies entirely within a JPEG block and the other borders or overlaps a JPEG block. A histogram of these differences is computed from all 8x8 non overlapping image blocks. A 8x8 “blocking artifact” matrix (BAM) is computed as the average difference between these histograms. For uncompressed images, this matrix is random, while for a compressed image, this matrix can show some patterns. Such patterns are often observed manually or identified using 
supervised pattern classification.
The main disadvantage was that format based methods were mostly restricted to JPEG compressed images. Their performance was not checked on other image formats. 


\subsection{Block-based method}
The block-based method mainly used to detect copy-move forgeries, where we select one window and compare in whole other image. The block-based matching can involve lexicographical matching, hashing, Euclidean distance as some of the approaches used to match the blocks based on defined thresholds, comparing the similarity between these blocks are time taken.\cite{inproceedings}

\subsection{Pixel based technique}
Some forgery detection approaches directly operate on pixels. For example, two computationally efficient algorithms have been developed to detect cloned image regions\cite{Popescu04exposingdigital,5999526}. In \cite{Popescu04exposingdigital} the author overcome the time complexity by projecting the image in to lower dimension using (PCA). Then they apply matching algorithms which reduce the computation like calculating euclidean distance.
Another common form of photographic manipulation is the digital splicing of two or more images into a single composite. When performed carefully, the border between the spliced regions can be visually imperceptible. 

In \cite{10.1007/978-3-540-87442-3_136,zhao2010detecting}, the authors show that splicing disrupts higher-order Fourier statistics, which can subsequently be used to detect the forgery. In image splicing sharp edges are usually introduced. To detect the image splicing forgery the author uses chrominance channel, and the forgeries are based on observation of the edge maps of these. A common limitation of most traditional methods, is that the judgement of the forgery has to be made manual observations.


\subsection{Learning based technique}
Among the learning based methods, some of the earlier methods use features which are extracted by using SIFT, SURF etc.\cite{4756779} and identify forged images  using SVM or deep neural network. However, in more recently, the popularity and performance deep learning methods has also proven to benefit the forgery detection area\cite{abdalla2019convolutional,kumar2020syn2real,rao2016deep,wang2020intelligent}. While in some earlier works on deep fakes interpretability also has been considered, this has been restricted for face images. Our work reported here is on similar lines but we consider forgeries in general natural images. 

\section{Proposed work} 
As indicated earlier, our focus on passive method is not just from the perspective of detection of copy-move and inpainting forgeries, but also on interpreting the classification via which one can also identifying the location of the forged regions. 

For this purpose, we use a forged dataset that we synthesize ourselves. 
We then use an in-house convolutional neural network (CNN) for classification, and finally employ Grad-CAM to interpret the results, which also yields approximate forgery locations. We describe these in the three subsections below:

\subsection{Dataset Generation}
For the forgery detection tasks, large natural image datasets required to train CNNs, are not available (except for face forgeries in the deepfake domain). 
Hence, we generate synthetic datasets for inpainting and copy-move forgery using COCO dataset. Fig. 1 shows some synthetically generated images. COCO (Common Object in Context) dataset is publicly available for multiple purpose tasks in computer vision such as object detection, semantic segmentation, and key-point detection \cite{lin2015microsoft}. To generate our forgery dataset, we use the pixel-level object mask information provided in the COCO data. 

\begin{figure}[h]
            \begin{tabular}{c|c|c}
                \textbf{Original} & \textbf{Copy move} & \textbf{Inpainting}\\   
            \includegraphics[width=4cm,height=2.5cm]{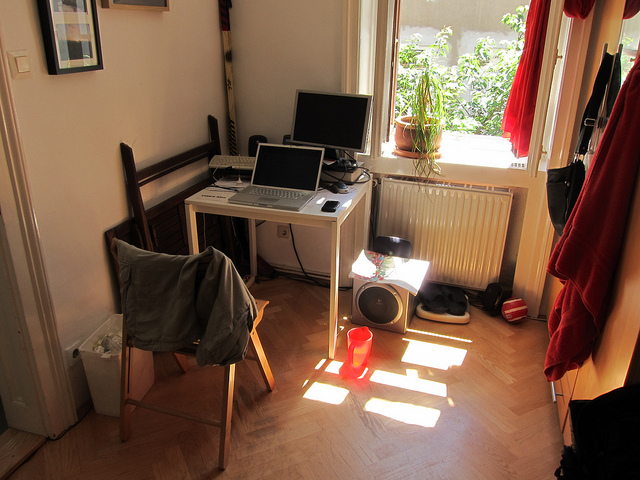}
            &\includegraphics[width=4cm,height=2.5cm]{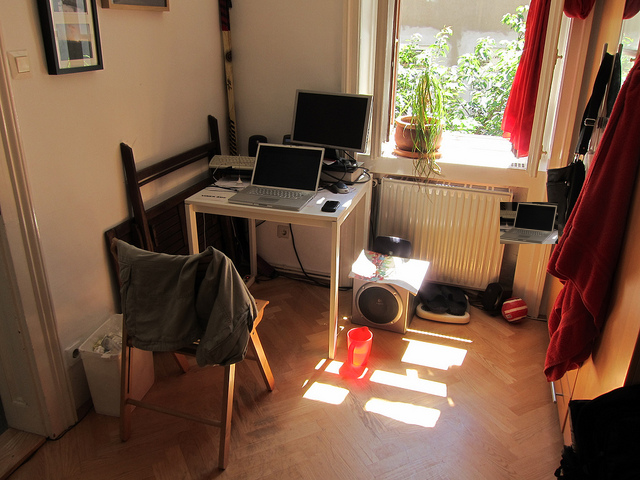}
            &\includegraphics[width=4cm,height=2.5cm]{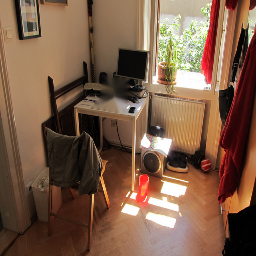}
             \\
            \includegraphics[width=4cm,height=2.5cm]{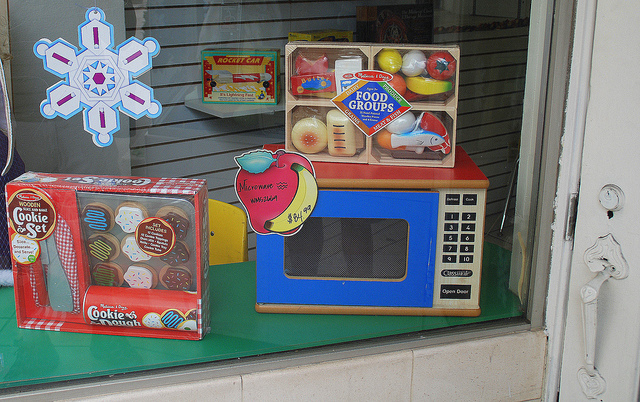}
            &\includegraphics[width=4cm,height=2.5cm]{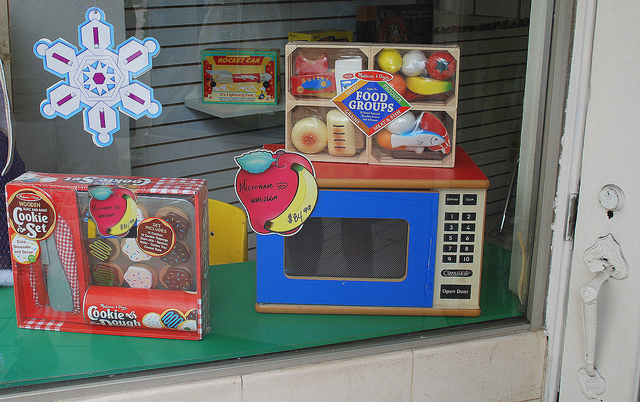}
            &\includegraphics[width=4cm,height=2.5cm]{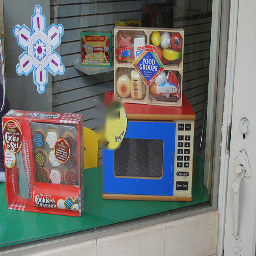}\\
            \includegraphics[width=4cm,height=2.5cm]{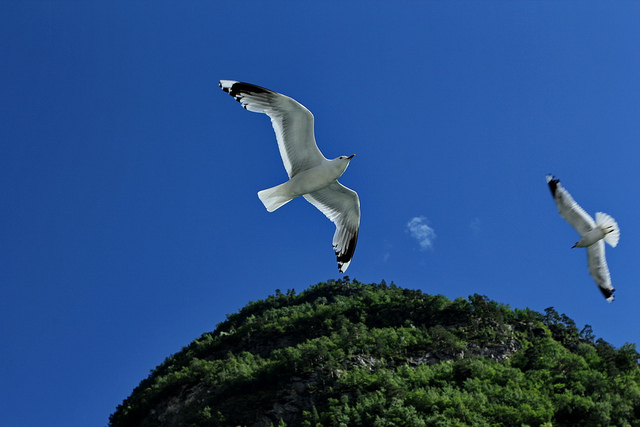}
            &\includegraphics[width=4cm,height=2.5cm]{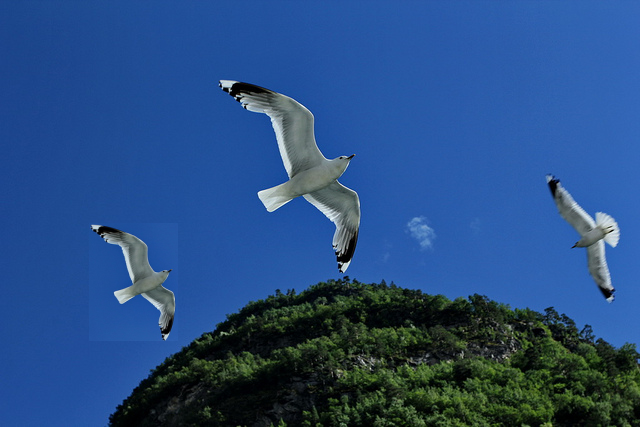}
            &\includegraphics[width=4cm,height=2.5cm]{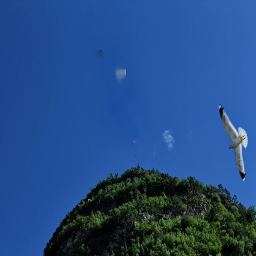}\\
            \includegraphics[width=4cm,height=2.5cm]{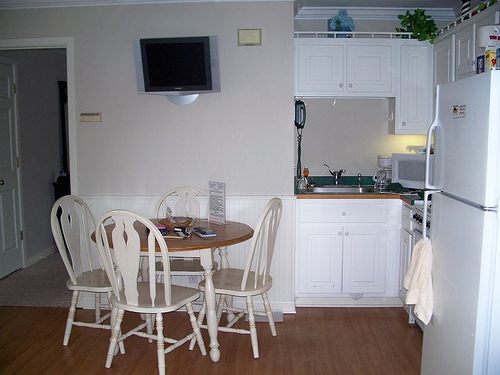}
            &\includegraphics[width=4cm,height=2.5cm]{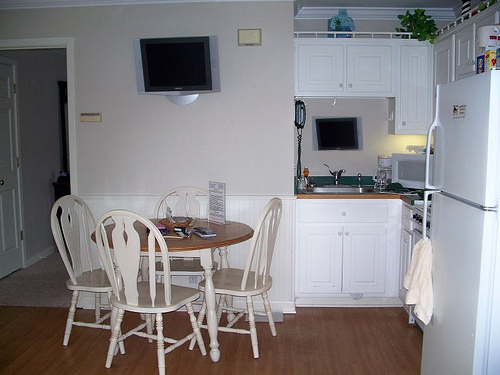}
            &\includegraphics[width=4cm,height=2.5cm]{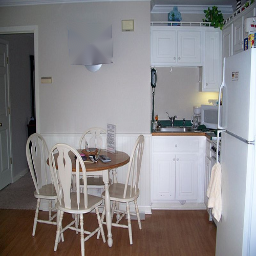}\\

            \textbf{a} & \textbf{b} & \textbf{c}
            \label{syndata}
             \end{tabular}
             \caption{a: Authentic image, b: Copy move forgery c: Image inpainting}
  \end{figure}
 

\subsubsection{Copy-Move Forgery:}
To generate copy-move forged images, we selected a specific categories of the COCO dataset. Then, we took into consideration of each mask’s area belonging to that category. Comparing all the mask areas, we select the mask with the largest area. Now, this copied area is pasted over the image after affine transformations and a image blending\cite{forte2020f} operation shown in Fig.2. 
$$ I_{f} = \alpha F + (1 - \alpha B)$$
where $I_{f}$ is final image, F is foreground object, B is background image and $\alpha$ is the blending factor.
Blending helps the image to fuse in another image smoothly and Deep Image Matting operation helps to fit on the second image. With this approach, we created a dataset of approximately 60,000 images.

\begin{figure}[h]
             \includegraphics[width=13cm,height=5.5cm]{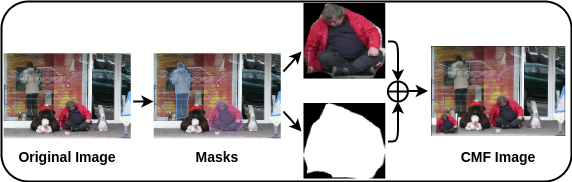}
             \caption{Method of Copy-move data generation}
            \label{cmf}
\end{figure}

\subsubsection{Semantic Inpainting}
We synthesized a dataset of inpainted images using all the sub-categories of the COCO dataset equally. The mask of particular sub-categories was cropped out. After that the Edge-connect inpainting method \cite{nazeri2019edgeconnect} is used. It is a type of deep Semantic inpainting that uses a two-stage approach to complete an image. Firstly, the edge generator fills out the missing edges, and then the image completion network completes the image based on the edges deduced as shown in Fig. 3. Using the above approach, we created approximately 21,000 inpainted images.

\begin{figure}[h]
             \includegraphics[width=13cm,height=5.5cm]{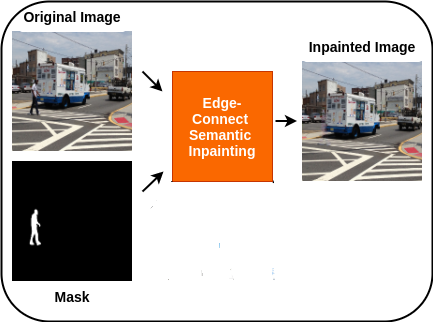}
             \caption{Method of Inpainting data generation}
            \label{fig:my_label}
  \end{figure}

\subsection{CNN Architecture}
For classification, we train three CNN models using three datasets as follows:\\
1. Model-1 (for Inpainting data only )\\
2. Model-2 (for Copy-Move data only)\\
3. Model-3 (for combined data [Inpainting + Copy-Move dataset])\\
We use five layer CNN architecture as a feature extractor having convolutions with pooling and batch norm layers. 
These are then followed by feature are further classified by fully connected layer into two classes (forged or non-forged class). We use the binary cross entropy loss, which is backpropagated using RMSProp optimizer. Fig. 4 shows the architecture of CNN.

\begin{figure}
            \includegraphics[width=14cm,height=6.5cm]{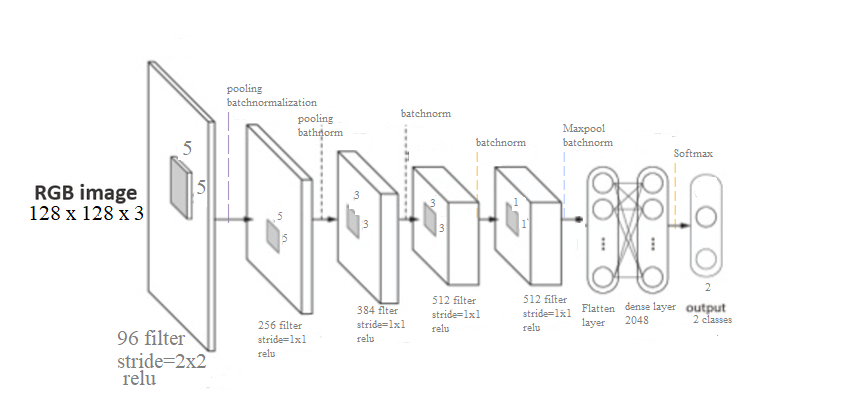}
            \caption{CNN architecture for classifying image forgery in to forge or non forge category.\\
            (Edited version of the original figure from \cite{unknown})}
            \label{fig:cnn}
 \end{figure}

\subsection{Grad-CAM}
After classifying the image in to forge or non-forge category, we employ the Grad-CAM method for interpreting the results. Specifically, the idea is to observe which image regions contribute to the classification, and ideally these should be the forged regions in cases of the forged images. For real images, this is not important, and hence we consider only the images classified as forged images for this analysis. We ensure that a reasonably good classification performance is achieved so that the interpretability can be relied upon. Grad-CAM yields a heatmap highlighting regions based on their importance to classification. We breifly describe the Grad-CAM method below. 

Grad-CAM stand for Gradient-Class Activation Mapping. It is a technique for producing ‘visual explanations’ for decisions from a large class of CNN Convolutional Neural Network based models. 
In (Grad-CAM) the gradients of any target concept are propagated to the final convolutional layer to produce a  localization map / heat map that  highlights the important area in the image for predicting the concept \cite{selvaraju2017grad}. To obtain the class-Activation map\cite{selvaraju2017grad}, Grad-CAM calculates the gradient of $y_c$ (score for class c) with respect to feature maps $A$ of a convolutional layer. these gradients flowing back are global-average-pooled to obtain the importance weights $\alpha^c_k$:

 $$\alpha^c_k = \frac{1}{z}\sum_i\sum_j\dfrac{\partial y^c}{\partial A^k_{ij}} $$
 finally Grad-CAM heat-map is a weighted combination of feature maps, followed by a ReLU activation function as shown in Fig. 5:
 
 $$L^C_{Grad-CAM} = ReLU (\sum_k \alpha^c_k A^K )$$\\
 \begin{figure}[h]
             \includegraphics[width=13cm,height=8cm]{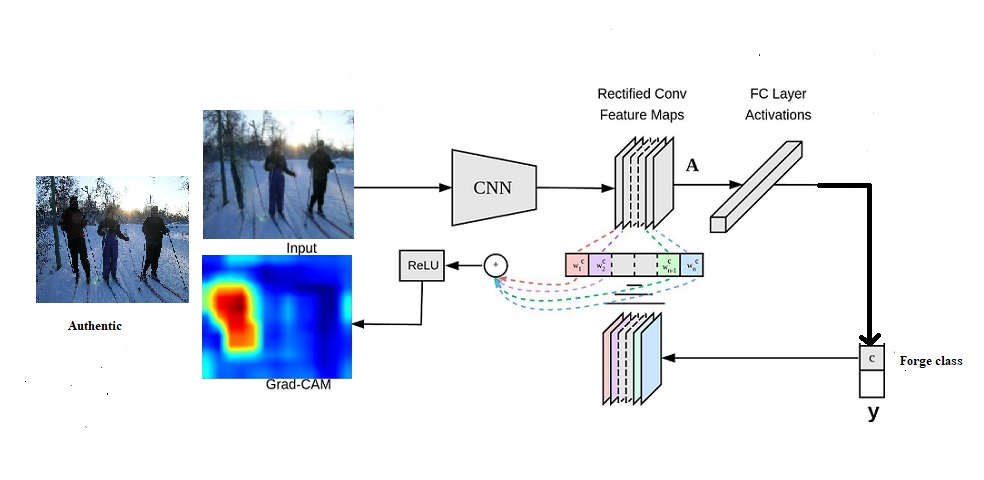}
             \caption{Grad-CAM Functionality (Edited version of the original figure from \cite{selvaraju2017grad}) }
            \label{fig:my_label}
  \end{figure}
  
\section{Experiment \& Results}
As indicated above, we train three CNN based model for three different scenarios, to assess results on individual forgery types, as well as for the combined data.
\subsubsection{Model-1 Trained on Inpainting dataset}
Inpainting dataset: (60,000 images, 75\%train data , 25\%validation data) we train the 60k images of inpainting forgery (30k authentic, 30k forged). we split the dataset 75\% as a training data and 25\% as a validation data. 
\subsubsection{Model-2 Trained on copy-Move dataset}
Copy-Move dataset: (60,000 images, 75\%train data , 25\%validation data)we train the 60k images of Copy-Move forgery (30k authentic,30k forged). we split the dataset 75\% as a training data and 25\% as a validation data. 
\subsubsection{Model-3 Trained on copy-Move + Inpainting dataset}
Model-3 trained on Copy-Move+Inpaint dataset. Note that this is a more realistic scenario, as in practical case, one would not know the forgery type a-priori.  

The classification results are shown in Table 1. We note that on the solo models the forgery detection in case of inpainting is better than that in case of copy-move forgery. This could be due to the fact that the network is able to learn subtle inpainting artifacts better as a complete region  is approximated with neighbourhood texture / colour. While the results for the copy-move case are also reasonably good, there are some misclassification. These can be attributed to the fact that the forged region is also natural-looking region from the image, and the artifacts would largely be affecting the borders. 

For the combined model, while we expect some drop in the classification performance, it is noted that the reduction is not significant in case of copy-move forgery and for the inpainting case, the combined model still yields well above 70$\%$. 


\begin{table}[htbp]
\caption{Classification results}
\begin{center}
\begin{tabular}{|c|c|c|c|}
\hline
\textbf{Test Dataset}&\multicolumn{3}{|c|}{\textbf{Testing Accuracy}} \\
\cline{2-4} 
\textbf{.} & \textbf{\textit{Model-1}}& \textbf{\textit{Model-2}}& \textbf{\textit{Model-3}} \\
\hline
Copy-Move dataset&NA&70\% & 69\%  \\
\hline
Inpainting dataset& 80\% &NA& 73\%  \\
\hline
Inpaint+CopyMove&NA&NA&70\%  \\
\hline
\end{tabular}
\label{tab1}
\end{center}
\end{table}

\subsection{Grad-CAM Results}
We now demonstrate some of the typical visual results for demonstrating the interpretability on all the three models.

As we observe in Fig. 6 the grad-cam results are satisfactory (even in cases where the inpainted regions are relatively small). The Grad-cam generates the heatmap in forged region (Inpainted area) well. This indicates that our CNN model (Model-1) learns right features from the input images and classify properly with valid forged features.
\begin{figure}[h]
            \begin{tabular}{|c|c|}
            \hline
          \textbf{original}&\textbf{forged image$~~~$heatmap$~~~$overlayheatmap}\\   
            \hline
         \includegraphics[width=3cm,height=3cm]{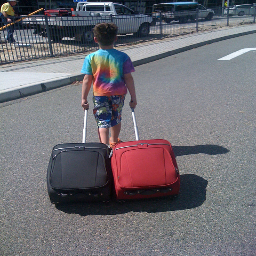}&\includegraphics[width=9cm,height=3cm]{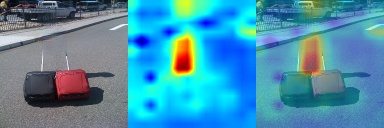}\\
          \includegraphics[width=3cm,height=3cm]{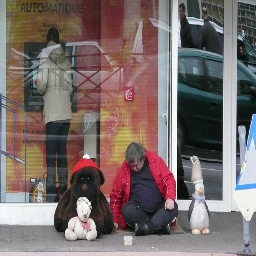}&\includegraphics[width=9cm,height=3cm]{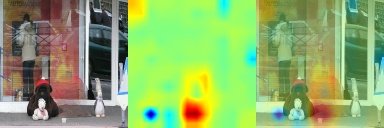}\\
             \includegraphics[width=3cm,height=3cm]{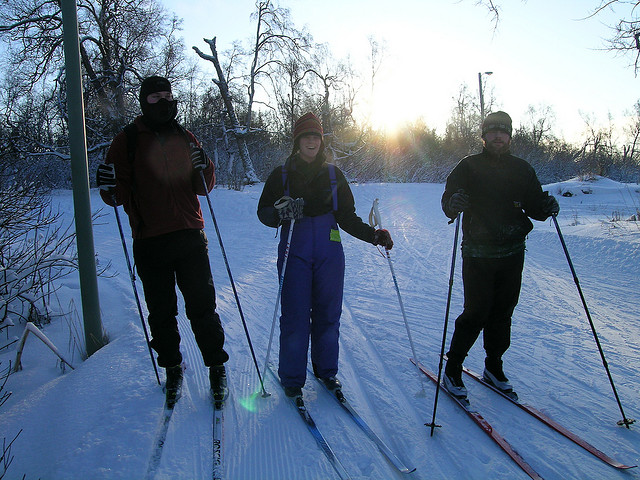}&\includegraphics[width=9cm,height=3cm]{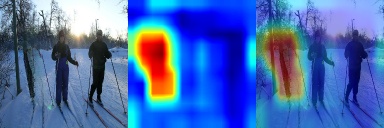}\\
            \hline
            \end{tabular}
             \caption{Above figure shows the grad-cam results of image inpainting using Model-1}
            \label{fig:my_label}
  \end{figure}
  
Fig. 7 shows the Grad-CAM results of copy-move dataset. As we can observe the  high intensity values of the heatmap (red) heatmap correctly superimposes on the copy-paste area. Interestingly the localization is quite accurate, in cases where the images are classified correctly. So, in this case too, we can conclude that our Model-2 learns suitable features from the input dataset.

\begin{figure}[h]
            \begin{tabular}{|c|c|}
            \hline
          \textbf{original}&\textbf{forged image$~~~$heatmap$~~~$overlayheatmap}\\   
            \hline
         \includegraphics[width=3cm,height=3cm]{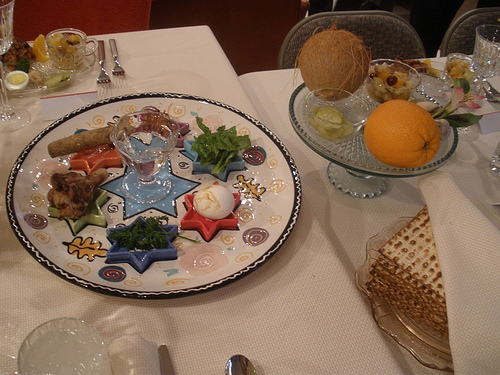}&\includegraphics[width=9cm,height=3cm]{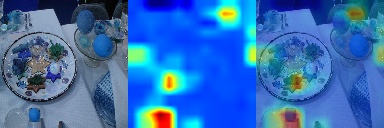}\\
          \includegraphics[width=3cm,height=3cm]{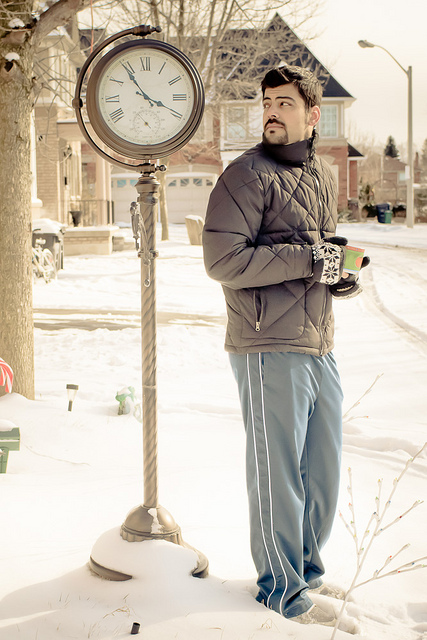}&\includegraphics[width=9cm,height=3cm]{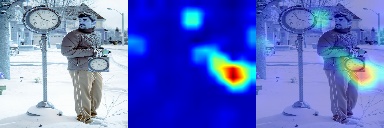}\\
         \includegraphics[width=3cm,height=3cm]{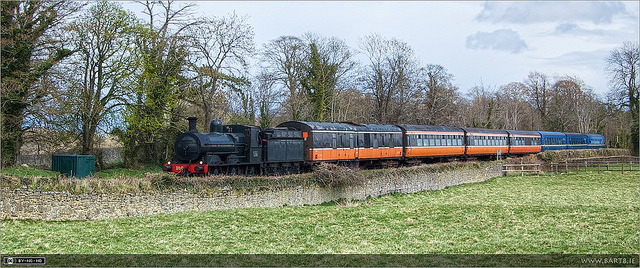}&\includegraphics[width=9cm,height=3cm]{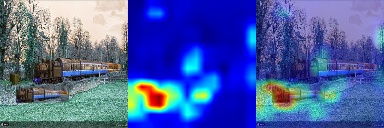}\\
            \hline
            \end{tabular}
             \caption{Above figure shows the grad-cam results of image copy-move using Model-2}
            \label{fig:my_label}
  \end{figure}
Finally, Fig. 8 shows the Grad-CAM results of copy-move dataset \& Inpainting dataset. In spite of the combined training, the Grad-CAM results demonstrate good explanabilty on the correctly classifed samples, irrespective of the kind of forgery. 

\begin{figure}[h]
            \begin{tabular}{|c|c|}
            \hline
          \textbf{original}&\textbf{forged image$~~~$heatmap$~~~$overlayheatmap}\\   
            \hline
         \includegraphics[width=3cm,height=3cm]{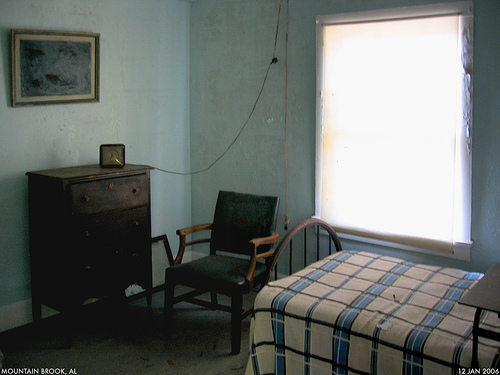}&\includegraphics[width=9cm,height=3cm]{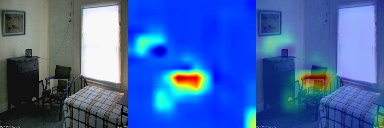}\\
          \includegraphics[width=3cm,height=3cm]{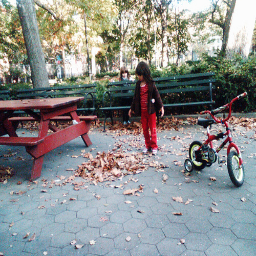}&\includegraphics[width=9cm,height=3cm]{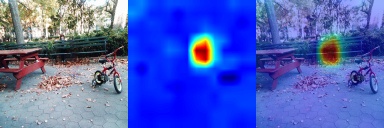}\\
        \includegraphics[width=3cm,height=3cm]{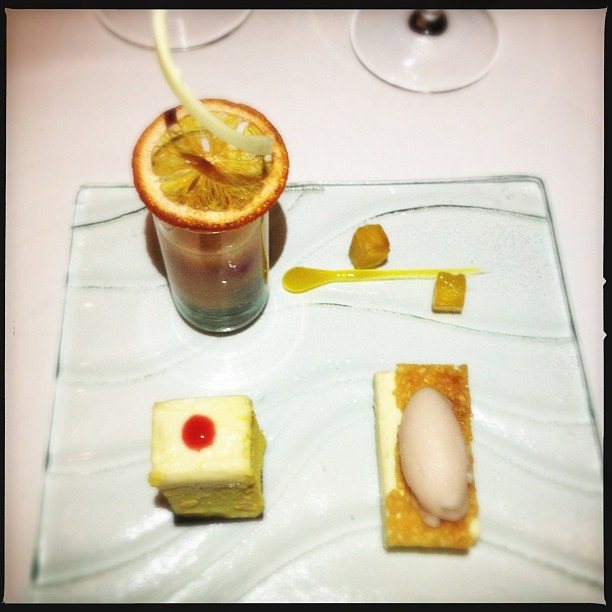}&\includegraphics[width=9cm,height=3cm]{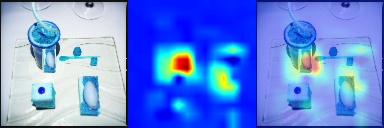}\\
            \hline
            \end{tabular}
             \caption{Above figure shows the grad-cam results of image inpainting + Copymove using Model-3}
            \label{fig:my_label}
  \end{figure}

\newpage
\section{Conclusion}
In this work, we demonstrated an interpretable approach to image forgery, focusing on two types of image forgeries, and across three CNN models. We show that the forgery detection can be explainable even for small forged regions from general natural images. While, our focus was on considering the explanability, and not merely on achieving very high results, we also show that a combined model learning two different types of forgeries can yield reasonably good results on random forgeries (unlike some recent work which consider face forgeries in deep-fakes). We note that the explanability analysis can be generalized to other networks which can yield better performance too, and should be an important consideration for image based classifiers for localizing forgeries. 
\bibliographystyle{IEEEtran}
\bibliography{THESIS}
 \end{document}